# Multimodal Sentiment Analysis: Perceived vs Induced Sentiments

Aditi Aggrawal
*Department of Information Technology*
*Delhi Technological University*
*Delhi, India*
aditiagg99@gmail.com

Dr. Deepika Varshney
*Department of Computer Science and Engineering*
*Jaypee Institute of Information Technology,*
*Delhi, India*
deepika.varshney@mail.jiit.ac.in

Saurabh Patel
*Department of Information Technology*
*Delhi Technological University*
*Delhi, India*
srbhptl39@gmail.com

*Abstract*—Social media has created a global network where people can easily access and exchange vast information. This information gives rise to a variety of opinions, reflecting both positive and negative viewpoints. GIFs stand out as a multimedia format offering a visually engaging way for users to communicate. In this research, we propose a multimodal framework that integrates visual and textual features to predict the GIF sentiment. It also incorporates attributes including face emotion detection and OCR generated captions to capture the semantic aspects of the GIF. The developed classifier achieves an accuracy of 82.7% on Twitter GIFs, which is an improvement over state-of-the-art models. Moreover, we have based our research on the ReactionGIF dataset, analysing the variance in sentiment perceived by the author and sentiment induced in the reader.

## I. INTRODUCTION

People's emotions online can be genuine reactions, while also being influenced by what they see and interact with, creating a blend of authentic and induced feelings. This dynamic interplay of emotions reflects the diverse nature of online interactions. For example, someone might get excited about a new product trending online, even if they weren't initially interested. The information being shared online has taken diverse forms, including text, audio, visuals, or their combinations. Animated GIFs constitute one such format that has recently gained prominence online. A study of 3.7 million users suggests that these brief, silent, often low-resolution video clips are 55% more captivating than any other media format on Twitter[1].

The automatic detection of human emotions is a crucial aspect in affective computing, with applications in varied domains like marketing, psychology, education and customer service [1]. Drawing from music and films, emotions in text can be categorized as **perceived emotions**: recognized by readers, and **induced emotions**: aroused in readers. This distinction, except by Buechel in his work [2], is largely absent in NLP literature. Induced emotions can be gathered through physiological measurements or self-reporting, which is challenging due to complexity, cost, unreliability, and scalability involved in the process. To overcome such challenges, we have based our research on the use of Reaction

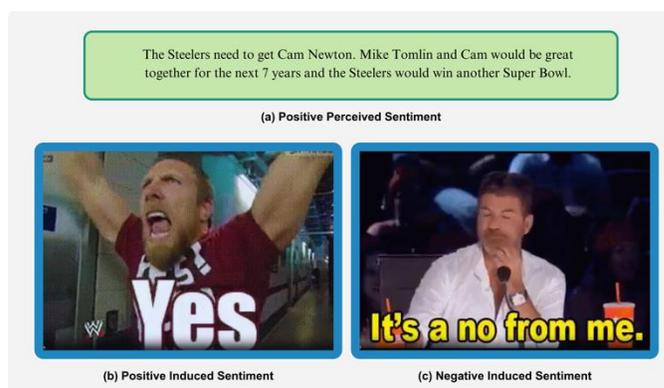

Figure 1: Example of Perceived vs induced sentiment

GIFs in online conversations, displaying "emotional responses to prior talk in text-mediated conversations" [3]. The ReactionGIF dataset [4] captures in-the-wild texts, inherently supervised with Reaction GIFs and their respective induced labels. Figure I describes an example of perceived sentiment of a tweet vs two opposing sentiments induced in readers.

Extensive research has been conducted on sentiment analysis, with applications ranging from forecasting political elections [5] to predicting box-office revenues [6] and economic indicators [7]. Most of the current sentiment analysis primarily focuses on textual data, however recently there has been a surge in popularity of visual content in the form of both images and videos. Notably, statistics show that memes and GIFs in tweets have significantly boosted user engagement[2], increasing clicks by 18%, retweets by 150%, and favorites by 89%. To the best of our knowledge, the implicated sentiment in GIFs has received limited attention. In our study, we examine the sentiments of popular GIFs accompanied by extraction of OCR generated short text captions and human face emotion detection. These texts serve as annotations for the GIFs, enabling sentiment analysis. Furthermore, the low-level statistics of GIFs differ from those of natural videos or images due to their small size and illustrative style.

Visual sentiment analysis involves complexities due to the high level of abstraction and subjectivity in human recognition, alongside various visual tasks like object and scene recognition.

---

[1]https://buffer.com/resources
[2]https://business.twitter.com/en/blog

The two main challenges posed when assessing sentiments in GIFs include understanding the video content and bridging the semantic gap between texts and videos [12]. First, video understanding often loses the sequence of images and videos in animated GIFs needed to extract its spatial-temporal features and to comprehend the GIFs' integrated sentiment content. Second, without semantic sentiment labels, the relationship of mid-level features from low-level features. As suggested in [13], lexicon-based methods base their overall sentiment predictions in textual messages on words annotated with polarity scores. However, Convolutional Neural Networks [8, 10, 11] have proven to be transformative for computer vision tasks, utilizing which some studies have made attempts to predict sentiment in images [9, 12] and videos [14]. Through this work, we propose a multi-modal framework that integrates cutting-edge techniques in both visual and textual analysis.

**Contributions** In this paper, we outline our contributions aimed at addressing the aforementioned challenges, as follows:

1. We are one of the first to study and analyse the distinction between perceived vs induced affective states among Twitter users, exploiting the online use of ReactionGIFs. We predict the perceived sentiment and induced sentiment from the original Tweet of the user and its ReactionGIF respectively.

2. We propose a multi-modal sentiment classification framework that effectively implements late fusion of visual-textual features to predict the final sentiment class of short animated GIFs circulated on social media. We leverage Optical Character Recognition and Face Emotion Recognition techniques to better capture the context of GIFs.

various table text styles are provided. The formatter will need to create these components, incorporating the applicable criteria that follow.

## II. RELATED WORK

In this section, we review existing research that closely aligns with our study on analyzing sentiments in social multimedia and on distinguishing between perceived and induced user sentiments.

### A. Identification of Affective States

The automatic detection of human affective states and the ability to distinguish between perceived emotion state and induced emotion state is a crucial aspect in affective computing. While well classified induced emotions data is significant for dialogues systems aiming to elicit specific emotional responses from users, most current emotion datasets are annotated with perceived emotions. Buechel et al. in [2] investigated how different perspectives (reader's, writer's, and text's) influence emotion annotations in written discourse. Their study further informed the creation of EMOBANK, a novel dataset for emotion analysis. Pool and Nissim [15] employed distant supervision approach and identified induced emotions in Facebook posts by analyzing six available emoji reactions. Despite its automatic nature, this method has limitations in emotional range, due to a small and fixed set of reactions, and in accuracy, due to issues arising from the visual ambiguity of emojis [16].

### B. Textual Sentiment Analysis

Traditional sentiment analysis in text focuses on lexicon-based features to extract embedded sentiment in the text. Methods like SentiWordNet [18] and VADER [17] use individual word scores and heuristic rules for sentence-level sentiment prediction. These interpretable and efficient tools are widely used in tasks such as consumer reviews and Twitter analysis. Recently, machine learning techniques like CNN and LSTM have been popular, with models like BERT dominating various NLP tasks. Both [19] and [20] employ pre-trained BERT to outperform previous state-of-the-art methods in sentiment analysis. [19] discusses the potential of using contextual word representations in complex aspect based sentiment analysis. In [20], Kaicheng et al, propose to adjust word weights based on the combined interaction of text and audio. Studies have explored aligning text genres between pre-training and testing, as seen in models like BERTweet [19], setting new benchmarks for semantic evaluation tasks in tweets and other social media text.he template is used to format your paper and style the text. All margins, column widths, line spaces, and text fonts are prescribed; please do not alter them. You may note peculiarities. For example, the head margin in this template measures proportionately more than is customary. This measurement and others are deliberate, using specifications that anticipate your paper as one part of the entire proceedings, and not as an independent document. Please do not revise any of the current designations.

### C. Visual Sentiment Analysis

For image sentiment analysis, earlier methods utilized low-to-mid-level features, such as color histograms [21] [22] and visual concepts in adjective-noun pairs (ANP) [22], combined with sentiment scores from textual metadata. With advancement in deep learning techniques, convolutional neural networks (CNNs), have been widely adopted for their expressive power. [23] and [25] explored transfer learning strategies on CNNs and based their work purely on visual features. Xu et al. [23] adopted CNN model pre-trained on ImageNet introducing a major advantage of not requiring any domain knowledge from linguistics or psychology. Song et al. [25] and You, Jin, and Luo [24] utilized attention modules in conjunction with CNN, to focus on salient image regions which reveal sentiment. Furthermore, to address the abstract and subjective nature of images, some studies [26][27] incorporated additional modules. Yuan et al. [26], employed facial expression recognition to enhance sentiment prediction for images with faces while [27] employed image caption modules to interpret objective image descriptions.

For video sentiment analysis, particularly in GIFs, researchers started with dataset collection and preparation. Li et al. [28] curated the Tumblr GIF (T-GIF) dataset, containing 100K animated GIFs and 120K natural language descriptions from Tumblr. In [29] and [30] experiments are conducted on the GIFGIF dataset from MIT Media Lab, which includes 6119 GIFs and 17 discrete emotion classes. Jou et al. [29] identified the most meaningful emotion-expressing features by comparing various features such as facial expressions, color histograms, and SB features in GIF videos. Building upon Jou's work, Chen [30] integrated short-term temporal sequence features extracted using 3D Convolutional Neural Networks (C3D) into the analysis. Also, [31] proposes a dataset and a semantic tree model based on spatial-temporal visual mid-level ontology.

Figure 2: Methodology

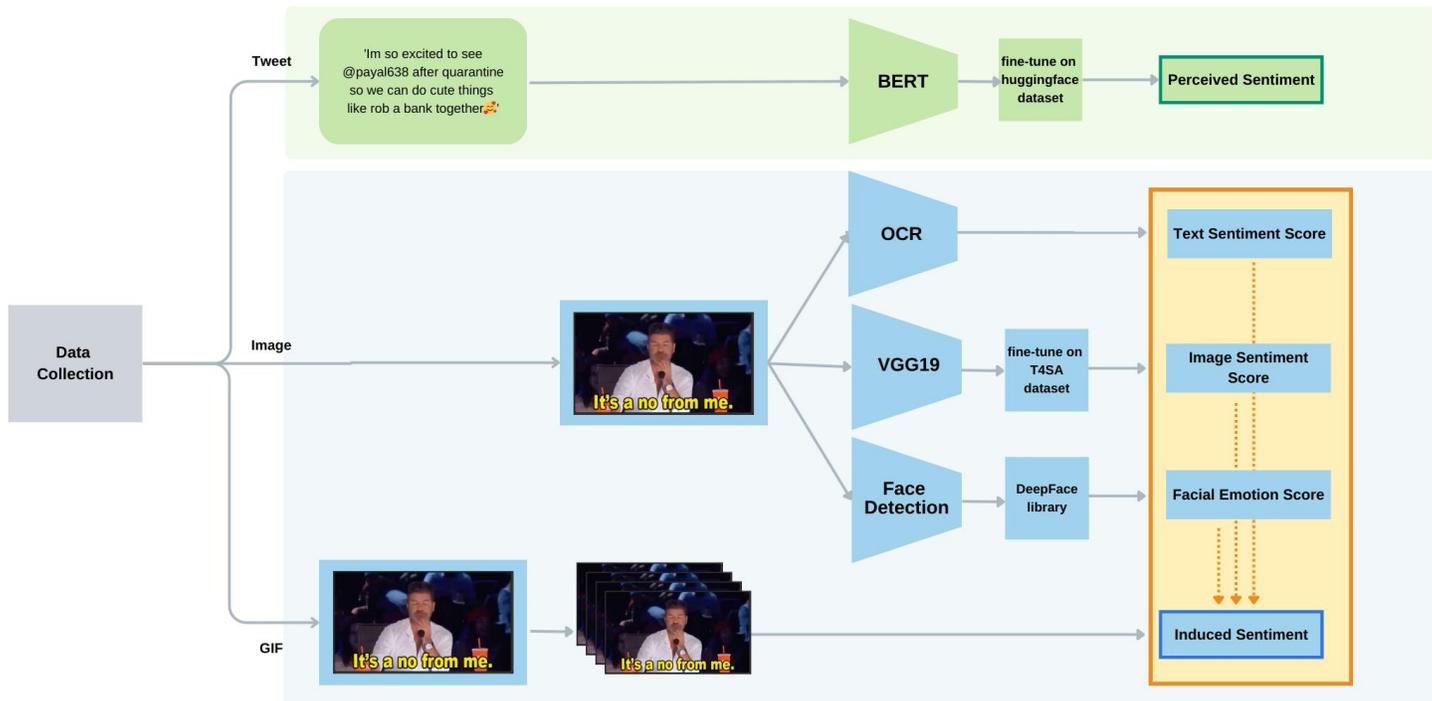

*D. Multimodal Sentiment Analysis*

As the name suggests this field focuses on combining information from both text and images to enhance sentiment prediction. With the field's growing popularity, three types of combination strategies have been explored for effective fusion: early fusion [32][33], intermediate fusion [12][34] and late fusion [35][36].

Early fusion combines multiple data sources directly into a single representation. [32] extracted linguistic, visual, and acoustic features to combine them for sentiment classification. Poria et al. [33] used deep convolutional neural networks to extract features from visual and textual modalities, fusing them with a multiple kernel learning classifier for sentiment analysis. However, early fusion methods can result in large, redundant input vectors and may not fully leverage modality correlation.

Intermediate fusion conducts the fusion process in intermediate layers of the neural networks. You et al. [12] proposed a cross-modality consistent regression model to ensure consistency between text and image representations. Another model by You et al. [34] integrated tree-structured LSTM with visual attention for image-text joint sentiment analysis. While achieving state-of-the-art performance, intermediate fusion-based methods may be affected when multimodal contents are incomplete.

Late fusion aggregates decisions from separate sentiment classifiers trained on distinct modalities. [35] and [36] employed late fusion for sentiment analysis using both text and images, combining n-gram textual features and mid-level visual features.

Late fusion assumes the modalities to be independent in the feature space, which may not hold true in practice due to high modality correlation.

III. PROPOSED WORK

Figure 2 illustrates our proposed framework. The system is divided into two training paths: first path for Perceived Sentiment Analysis performed on textual data and the second path for Induced Sentiment Analysis on video data. We have considered three possible modalities: textual (tweet), image and video (GIF). In the following sections, we elaborate on the details of our framework.

*A. Dataset*

We have performed our research on the publicly available ReactionGIF dataset, yielding 30k English language tweets with their reaction GIFs. This is a first-of-its-kind dataset focusing on two-turn conversations where each entry comprises a reaction GIF reply associated with a purely textual root post. Each of these tweets has been labeled with its reaction category. This category, conveying a rick affective signal, has been used to assign a suitable sentiment and emotion label to the entry, based on novel reaction-to-emotions mapping. We used Tweepy to fetch the tweets, downloaded the reaction GIFs and collected their associated metadata from the dataset.

*B. Textual Sentiment Analysis*

We have used pre-trained BERT model to predict the sentiment label for tweets and analyse perceived sentiments of the Twitter user. As a first step, data was cleaned using tweet-preprocessor library, insignificant characters and punctuations were removed, and the text was converted to lowercase in order to focus on the essential information. BERT is a deep learning model designed to capture relationships between words in a sentence and understand their context. Its unique ability to understand words in different contexts and generate embeddings based on the surrounding words, sets it apart from traditional text classification models. Text classification using BERT typically involves several key phases, as detailed below:

**Tokenization**

This phase involves conversion of raw text into a format that can be further processed by the model. BERT uses WordPiece tokenization, which breaks words into smaller meaningful subwords or characters. This method helps BERT handle rare words and morphemes effectively. This method helps BERT handle rare words and morphemes effectively. We have utilized the uncased-BERTBASE version in our work to convert our input text into token sequences.

**Model Architecture Configuration**

To enhance the performance of this model with 12 transformer blocks, we froze the embedding layer parameters. We adjusted the learning rate of encoder layers and the rest of them to 0.01 and 2e-5 respectively, so as to avoid overfitting. Our model was trained for 3 epochs, with batch size and maximum sequence length configured to 32 and 50 respectively. Additionally, we employed the Adam optimizer along with the MSE loss function.

**Fine-tuning**

This phase involves updating the pre-trained model's weights using backpropagation and task-specific labeled data. We have utilized huggingface pytorch implementation of BertForSequenceClassification to train our model on the huggingface dataset.

*C. Image Sentiment Analysis*

We have implemented VGG19 architecture to learn and extract visual features from raw image data. VGG19 is a 19 layers deep CNN architecture pretrained on ImageNet database, learning feature representations from a diverse set of images. In the ImageNet Large Scale Visual Recognition Challenge (ILSVRC) 2014, VGG19 achieved top-5 test accuracy of around 92.7%. VGG19 performs better in tasks requiring fine-grained object recognition due to its increased depth with 143.67 million parameters and ability to learn intricate representations.

To repurpose the VGG19 model from object detection to fit our usecase of sentiment analysis, we referred to [24]. We first freeze the convolutional base layers of the pretrained VGG19 model and then on top of them, we stack our custom layers utilizing global average pooling and softmax activation function to make a probability prediction. We compiled our model with categorical cross entropy loss and Adam optimizer with a learning rate of 10-4. Further, in order to finetune VGG19 for our research, we trained the configured model on the Twitter for Sentiment Analysis (T4SA) Dataset [37], setting the input image size to 48 x 48 pixels for all RGB images with a validation split of 0.2. We trained our model for a total of 3 epochs and a batch size of 32.

The T4SA dataset is a publicly available collection of 1.5M images. The researchers gathered both text and multimedia data from Twitter exhausting Twitter's API over a period of six months to gather about 1% of all tweets globally. To ensure the data was relevant for their sentiment analysis task, they applied specific rules discarding retweets, tweets without images and non-English tweets. After sentiment classification of the texts, they selected the tweets with the most confident sentiment predictions to create this dataset. The use of this dataset seems extremely relevant for our research due to its varied collection of images along with their source as Twitter.

*D. GIF Sentiment Analysis*

We have utilized the previously discussed image module sentiment analysis to determine the sentiment score of a GIF video file. Furthermore, our research proposes a novel technique incorporating face detection, emotion recognition and optical character recognition to enhance the average score obtained from the image module. We elaborate our approach within each individual technique below.

**Image Module**

A GIF video is split into frames, each of which is fed as a raw image to the aforementioned VGG19 model to generate a sentiment label. The available GIF videos are stored in '.mp4' format with their respective frames downloaded in '.jpg' format. One can either extract frames from a video at a constant frame count or a constant time period. Constant frame count is preferred to maintain consistent input size for each video clip while constant time period is often chosen to ensure uniform temporal coverage. Since the sentiment prediction of each GIF is independent, we went ahead with a constant time period of 0.1s to maintain consistent representation of the video's content. The final sentiment score is calculated as the average of these individual frame scores.

**Facial Emotion Detection**

Human facial expressions convey a range of emotions and analyzing these facial cues can provide valuable insights into a person's emotional state. We have utilized this technique of facial emotion recognition to determine the face emotion score of the GIF. We classified the GIF videos based on human face detection in their frames followed by prediction of the face emotions for videos where human face could be detected. We employed the DeepFace python framework for face detection and facial emotion analysis on image modules.

Deepface is a Python framework designed for face recognition and facial attribute analysis (age, gender, emotion and race). It combines powerful models like Google FaceNet, ArcFace, VGG-Face, Facebook Deepface, and others, with individual accuracies of 99.65%, 99.41%, 98.78%, 97.35% on the Labeled Faces in the Wild (LFW) dataset [38], the de facto benchmark dataset for studying unconstrained face recognition scenarios. The dataset contains labeled face photographs captured with variations in factors like pose, lighting, race, accessories, occlusions, and background, reflecting the natural variability encountered in real-world situations. Deepface handles the five common stages of a modern face recognition pipeline in the background: detect, align, normalize, represent and verify.

TABLE I. ATTRIBUTE ANALYSIS

| Face Detected? | OCR Caption Generated? | Percentage in Dataset | Accuracy |
|---|---|---|---|
| Yes | Yes | 48 | 0.827 |
| Yes | No | 23 | 0.723 |
| No | Yes | 17 | 0.761 |
| No | No | 12 | 0.550 |

We used the RetinaFace algorithm to detect human faces from the frames generated in GIF videos. RetinaFace is a single-stage designed for accurate face localization in real-world conditions, achieving an average precision of 91.4% on the WIDERFACE [39] hard test set. Then we predict and analyze emotions in the videos with human faces. The output of the DeepFace library for emotion recognition typically provides probabilities for a given facial image corresponding to six basic emotions. Of these, we labeled Happy, Surprise as emotions with positive sentiment while Sad, Angry, Fear, Disgust as emotions with negative sentiment. We calculate facial emotion score of a GIF video as the average predicted probability of emotions with positive and negative sentiments.

**Optical Character Recognition**

A caption inside a GIF can provide additional context, explaining the emotion or situation depicted in the GIF. We have utilized the Optical Character Recognition (OCR) technique to extract textual content from the frames of GIF videos. OCR technology analyzes the shapes, patterns, and features of characters in the scanned or photographed image and translates them into machine-readable text. This enables machines to automatically recognize characters, similar to how humans perceive objects through their eyes. This process mainly comprises the stages of text detection, text extraction and text recognition. The frames were first converted to grayscale to eliminate color complexities and then implemented OCR with an English dictionary based Tesseract engine. The generated OCR captions were analysed for their sentiment probability score exploiting the same BERT model fine-tuned during Text Sentiment Analysis.

Figure 3: Distribution in Sentiment Categories

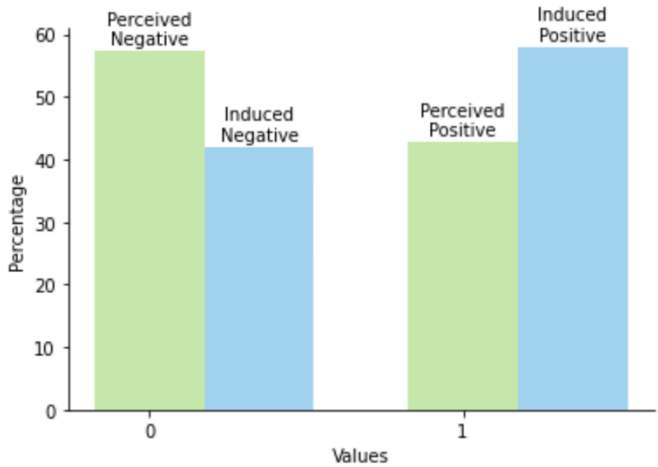

## IV. EXPERIMENTAL RESULTS

In this section, we elaborate on the performance of our methodology on the 25,774 entries of ReactionGIF dataset. We analyse and compare the sentiments of each entry from the perspective of the tweet's author with that of it's reader. We finally employ our multimodal framework to predict a combined sentiment of the input GIF Dataset.

### A. Perceived Sentiment vs Induced Sentiment

We define Perceived Sentiment as the sentiment recognized by the reader in the tweet, or the sentiment with which the user posted a tweet. Induced Sentiment is the sentiment invoked in the reader when he replies to the tweet through a GIF. Figure 3 displays the percentage distribution of these sentiments in the dataset, based on our predictions of tweet sentiment. We have utilised the aforementioned BERT model to predict perceived sentiment class of the tweets and have considered the sentiment label of reactionGIF in the dataset as the induced sentiment. The dataset comprises 17,714 distinct tweets and 18,976 unique GIFs. A tweet can have multiple replies with differing reader perspectives and hence multiple reactionGIFs, however, the dataset is biased towards tweets with a single reactionGIF with 15,731 (88%) such entries. There is an outlier tweet with 330 different reactionGIFs.

TABLE II. COMPARISON OF IMAGE SENTIMENT CLASSIFICATION

| Face Detected? | OCR Caption Generated? | Percentage in Dataset | Accuracy |
|---|---|---|---|
| Yes | Yes | 48 | 0.827 |
| Yes | No | 23 | 0.723 |
| No | Yes | 17 | 0.761 |
| No | No | 12 | 0.550 |

Moreover, Figure 4 showcases the distribution of instances across four possible combinations with: 1) same Perceived and Induced sentiments, and 2) opposing Perceived and Induced sentiments.

Figure 4: Distribution in Sentiment Combinations

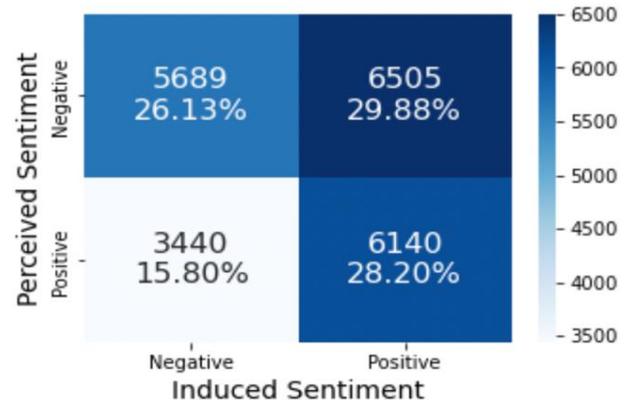

*B. Multimodal Sentiment Classification Framework*

To determine the GIF sentiment, we combined the predicted sentiment score from each of the three modules: Image Module, Face Detection Module, and OCR Module. We adopted late fusion technique to incorporate diverse features from different modalities. Each of the three modules predicts a probability sentiment score and we use the averaging fusion mechanism to generate the final sentiment class (0 - negative sentiment; 1 - positive sentiment). Table I states the composition of our dataset across two attribute classes: Face Detection and OCR generated caption. It also specifies the accuracy of our model in each of these cases. Additionally, our model's performance is further compared with existing models in Table II.

We conducted experiments using various datasets to fine-tune our pre-trained textual and visual models. The relevance of the huggingface dataset for the textual modality and the T4SA dataset for the visual modality is justified because both datasets are derived from the Twitter platform and provide a large corpus that encompasses diverse input media prevalent on the platform.

## V. CONCLUSION AND FUTURE SCOPE

Sentiment analysis of social multimedia, especially visual sentiment analysis, poses challenges in understanding diverse user behaviors commonly observed on social media. Additionally, it involves interpreting the abstraction and subjectivity inherent in images. This paper leverages deep learning to introduce a multimodal framework that integrates the textual and visual features into a unified model for predicting the GIF sentiment. The experimental results indicate that appropriately trained and fine-tuned convolutional neural networks outperform classifiers using low-level or mid-level visual features in image sentiment analysis. Our proposed model provides a more context aware sentiment prediction of the GIFs and significantly improves the performance by taking into account attributes such as face emotion detection and caption generation. Furthermore, our examination of the variation in author and reader perspectives for the same tweet reveals a close ratio (1.22) between same perceived and induced sentiments, and opposing perceived and induced sentiments. We hope that our findings in sentiment analysis can serve as a catalyst for additional research on user generated content in the online domain.

We believe there are areas for improvement where our research can serve as a stepping stone for machines to further enhance their performance in this domain. 1) Fine-grained emotion detection: We have focused our research on binary classification which can be extended to predict multiclass emotions. 2) Semantic visual understanding: Similar to face emotion detection and OCR caption generation, attributes such as the identification of objects and animals, image caption generation can help in an even better semantic understanding of the images. 3) Context-based predictions: Since the same GIF can be used for multiple expressions, models can be developed to make context aware predictions for a GIF by considering the conversation around it. For instance, a clapping hands GIF can be used for multiple expressions such as, sarcasm or applaud.